\documentclass[11pt,a4paper]{article}
\usepackage[hyperref]{acl2020}
\usepackage{times}
\usepackage{latexsym}

\usepackage{microtype}

\aclfinalcopy 

\usepackage{amsmath}
\usepackage{adjustbox}
\usepackage{comment}
\usepackage{soul}

\title{Evaluating Robustness to Input Perturbations\\for Neural Machine Translation}
\author{Xing Niu, Prashant Mathur, Georgiana Dinu, Yaser Al-Onaizan \\
Amazon AI \\
\texttt{\{xingniu,pramathu,gddinu,onaizan\}@amazon.com} \\}

\begin{document}

\maketitle

\begin{abstract}
    Neural Machine Translation (NMT) models are sensitive to small perturbations in the input. Robustness to such perturbations is typically measured using translation quality metrics such as BLEU on the noisy input. This paper proposes additional metrics which measure the relative degradation and changes in translation when small perturbations are added to the input. We focus on a class of models employing subword regularization to address robustness and perform extensive evaluations of these models using the robustness measures proposed. Results show that our proposed metrics reveal a clear trend of improved robustness to perturbations when subword regularization methods are used.
\end{abstract}

\section{Introduction}
Recent work has pointed out the challenges in building robust neural network models \citep{43405, DBLP:journals/corr/PapernotMGJCS16}. For Neural Machine Translation (NMT) in particular, it has been shown that NMT models are brittle to small perturbations in the input, both when these perturbations are synthetically created or generated to mimic real data noise \citep{BelinkovB18}. Consider the example in Table~\ref{tab:example} where an NMT model generates a worse translation as a consequence of only one character changing in the input.

Improving robustness in NMT has received a lot of attention lately with data augmentation
\citep{sperber-etal-2017-toward,BelinkovB18,vaibhav-etal-2019-improving,liu-etal-2019-robust,karpukhin2019} and adversarial training methods \citep{cheng2018,ebrahimi-etal-2018-adversarial,cheng-etal-2019-robust,michel-etal-2019-evaluation} as some of the more popular approaches used to increase robustness in neural network models.

In this paper, we focus on one class of methods, subword regularization, which addresses NMT robustness without introducing any changes to the architectures or to the training regime, solely through dynamic segmentation of input into subwords \citep{kudo-2018-subword,bpe-dropout}. We provide a comprehensive comparison of these methods on several language pairs and under different noise conditions on robustness-focused metrics.

\begin{table}[t]
\centering
\adjustbox{width=\columnwidth}{
\begin{tabular}{l|l}
    Original input & Se kyll\"a tuntuu sangen luultavalta. \\
    Translation & It certainly seems very likely. \\
    \hline
    Perturbed input & Se kyll\"a tu\hl{m}tuu sangen luultavalta. \\
    Translation & It will probably darken quite probably. \\
    \hline
    Reference & It certainly seems probable. \\
\end{tabular}}
\caption{An example of NMT English translations for a Finish input and its one-letter misspelled version.}
\label{tab:example}
\end{table}

Previous work has used translation quality measures such as BLEU on noisy input as an indicator of robustness. Absolute model performance on noisy input is important, and we believe this is an appropriate measure for noisy domain evaluation \citep{michel-neubig-2018-mtnt,berard-etal-2019-machine,li-etal-2019-findings}. However, it does not disentangle model quality from the relative degradation under added noise.

For this reason, we propose two additional measures for robustness which quantify the changes in translation when perturbations are added to the input. The first one measures relative changes in translation quality while the second one focuses on consistency in translation output irrespective of reference translations. Unlike the use of BLEU scores alone, the metrics introduced show clearer trends across all languages tested: NMT models are more robust to perturbations when subword regularization is employed. We also show that for the models used, changes in output strongly correlate with decreased quality and the consistency measure alone can be used as a robustness proxy in the absence of reference data.
\section{Evaluation Metrics}

Robustness is usually measured with respect to translation quality. Suppose an NMT model $M$ translates input $x$ to $y'$ and translates its perturbed version $x_\delta$ to $y'_\delta$, the translation quality (TQ) on these datasets is measured against reference translations $y$: $\texttt{TQ}(y',y)$ and $\texttt{TQ}(y'_\delta,y)$. TQ can be implemented as any quality measurement metric, such as BLEU \citep{papineni-etal-2002-bleu} or 1 minus TER \citep{snover2006study}.

Previous work has used TQ on perturbed or noisy input as an indicator of robustness. However, we argue that assessing models' performance relative to that of the original dataset is important as well in order to capture models' sensitivity to perturbations. Consider the following hypothetical example:
\begin{align*}
    M_1\text{: } & \texttt{BLEU}(y'_1,y) = 40, \texttt{BLEU}(y'_{\delta1},y) = 38;\\
    M_2\text{: } & \texttt{BLEU}(y'_2,y) = 37, \texttt{BLEU}(y'_{\delta2},y) = 37.
\end{align*}
Selecting $M_1$ to translate noisy data alone is preferable, since $M_1$ outperforms $M_2$ ($38>37$). However, $M_1$'s quality degradation ($40\rightarrow38$) reflects that it is in fact more sensitive to perturbation $\delta$ comparing with $M_2$.

To this end, we use the ratio between $\texttt{TQ}(y',y)$ and $\texttt{TQ}(y'_\delta,y)$ to quantify an NMT model $M$'s invariance to specific data and perturbation, and define it as \textbf{robustness}:
\begin{align*}
    \texttt{ROBUST}(M|x,y,\delta) = \frac{\texttt{TQ}(y'_\delta,y)}{\texttt{TQ}(y',y)}.
\end{align*}
When evaluating on the dataset $(x,y)$, $\texttt{ROBUST}(M|x,y,\delta)<1$ means the translation quality of $M$ is degraded under perturbation $\delta$; $\texttt{ROBUST}(M|x,y,\delta)=1$ indicates that $M$ is robust to perturbation $\delta$.

It is worth noting that: (1) $\texttt{ROBUST}$ can be viewed as the normalized $\Delta\texttt{TQ}=\texttt{TQ}(y',y)-\texttt{TQ}(y'_\delta,y)$ because $\Delta\texttt{TQ}/\texttt{TQ}(y',y)=1-\texttt{ROBUST}$. We opt for the ratio definition because it is on a $[0,1]$ scale, and it is easier to interpret than $\Delta\texttt{TQ}$ since the latter needs to be interpreted in the context of the TQ score. (2) High robustness can only be expected under low levels of noise, as it is not realistic for a model to recover from extreme perturbations.

\paragraph{Evaluation without References}

Reference translations are not readily available in some cases, such as when evaluating on a new domain. Inspired by unsupervised consistency training \cite{xie2019}, we test if translation \textit{consistency} can be used to estimate robustness against noise perturbations. Specifically, a model is consistent under a perturbation $\delta$ if the two translations, $y'_\delta$ and $y'$ are similar to each other. Note that consistency is sufficient but not necessary for robustness: a good translation can be expressed in diverse ways, which leads to high robustness but low consistency.

We define \textbf{consistency} by
\begin{align*}
\texttt{CONSIS}(M|x,\delta) = \texttt{Sim}(y'_\delta,y').
\end{align*}
\texttt{Sim} can be any \textit{symmetric} measure of similarity, and in this paper we opt for $\texttt{Sim}(y'_\delta,y')$ to be the harmonic mean of $\texttt{TQ}(y'_\delta,y')$ and $\texttt{TQ}(y',y'_\delta)$, where \texttt{TQ} is BLEU between two outputs.

\section{Experimental Set-Up}

We run several experiments across different language families with varying difficulties, across different training data conditions (i.e. with different training data sizes) and evaluate how different subword segmentation strategies performs across noisy domains and noise types. 

\paragraph{Implementation Details} We build NMT models with the Transformer-base architecture \citep{VaswaniSPUJGKP17} implemented in the Sockeye toolkit \citep{sockeye}. The target embeddings and the output layer’s weight matrix are tied \citep{press-wolf-2017-using}. Training is done on 2 GPUs, with a batch size of 3072 tokens and we checkpoint the model every 4000 updates. The learning rate is initialized to 0.0002 and reduced by 10\% after 4 checkpoints without improvement of perplexity on the development set. Training stops after 10 checkpoints without improvement.

\paragraph{Tasks and Data} We train NMT models on eight translation directions and measure robustness and consistency for them. \texttt{EN$\leftrightarrow$DE} and \texttt{EN$\leftrightarrow$FI} models are trained with pre-processed WMT18 news data and tested with the latest news test sets (newstest2019).

Recently, two datasets were built from user-generated content, MTNT \citep{michel-neubig-2018-mtnt} and 4SQ \citep{berard-etal-2019-machine}. They provide naturally occurring noisy inputs and translations for \texttt{EN$\leftrightarrow$FR} and \texttt{EN$\leftrightarrow$JA}, thus enabling automatic evaluations. \texttt{EN$\leftrightarrow$JA} baseline models are trained and also tested with aggregated data provided by MTNT, i.e., KFTT+TED+JESC (KTJ). \texttt{EN$\leftrightarrow$FR} baseline models are trained with aggregated data of Europarl-v7 \citep{koehn2005europarl}, NewsCommentary-v14 \citep{bojar-etal-2018-findings}, OpenSubtitles-v2018 \citep{lison-tiedemann-2016-opensubtitles2016}, and ParaCrawl-v5\footnote{\url{https://paracrawl.eu/}},
which simulates the UGC training corpus used in 4SQ benchmarks, and they are tested with the latest WMT new test sets supporting \texttt{EN$\leftrightarrow$FR} (newstest2014).

Following the convention, we also evaluate models directly on noisy MTNT (mtnt2019) and 4SQ test sets. We fine-tune baseline models with corresponding MTNT/4SQ training data, inheriting all hyper-parameters except the checkpoint interval which is re-set to 100 updates. Table~\ref{tab:data} shows itemized training data statistics after pre-processing.

\begin{table}[t]
\centering
\adjustbox{width=\columnwidth}{
\begin{tabular}{l|l|r|r}
    & Languages & \# sentences & \# \texttt{EN} tokens \\
    \hline
    & \texttt{EN$\leftrightarrow$DE} & 29.3 M & 591 M \\
    BASE & \texttt{EN$\leftrightarrow$FR} & 22.2 M & 437 M \\
    & \texttt{EN$\leftrightarrow$FI} & 2.9 M & 71 M \\
    & \texttt{EN$\leftrightarrow$JA} & 3.9 M & 43 M \\
    \hline
    & \texttt{EN$\rightarrow$FR} & 36.1 K & 1,011 K \\
    MTNT & \texttt{FR$\rightarrow$EN} & 19.2 K & 779 K \\
    & \texttt{EN$\rightarrow$JA} & 5.8 K & 338 K \\
    & \texttt{JA$\rightarrow$EN} & 6.5 K & 156 K \\
    \hline
    4SQ & \texttt{FR$\rightarrow$EN} & 12.1 K & 141 K \\
\end{tabular}}
\caption{Statistics of various training data sets.}
\label{tab:data}
\end{table}

\paragraph{Perturbations} We investigate two frequently used types of perturbations and apply them to WMT and KTJ test data. The first is synthetic misspelling: each word is misspelled with probability of 0.1, and the strategy is randomly chosen from single-character deletion, insertion, and substitution \citep{karpukhin2019}. The second perturbation is letter case changing: each sentence is modified with probability of 0.5, and the strategy is randomly chosen from upper-casing all letters, lower-casing all letters, and title-casing all words \citep{berard-etal-2019-machine}.\footnote{Character substitution uses neighbor letters on the QWERTY keyboard, so accented characters are not substituted. Japanese is ``misspelled" for each character with probability of 0.1, and it only supports deletion and repetition. Letter case changing does not apply to Japanese.}

Since we change the letter case in the test data, we always report case-insensitive BLEU with `13a' tokenization using sacreBLEU \citep{post-2018-call}. Japanese output is pre-segmented with Kytea before running sacreBLEU.\footnote{\url{http://www.phontron.com/kytea/}}

\paragraph{Model Variations} We focus on comparing different (stochastic) subword segmentation strategies: BPE \citep{sennrich-etal-2016-neural}, BPE-Dropout \citep{bpe-dropout}, and SentencePiece \citep{kudo-2018-subword}. Subword regularization methods (i.e., BPE-Dropout and SentencePiece) generate various segmentations for the same word, so the resulting NMT model better learns the meaning of less frequent subwords and should be more robust to noise that yields unusual subword combinations, such as misspelling. We use them only in offline training data pre-processing steps, which requires no modification to the NMT model.\footnote{We sample one subword segmentation for each source sequence with SentencePiece.}

\section{Experimental Results}

\begin{table*}[!p]
\centering
\adjustbox{width=0.9\width}{
\begin{tabular}{l|l|rrr|rrr}
    & Model & BLEU & \texttt{ROBUST} & \texttt{CONSIS} & BLEU & \texttt{ROBUST} & \texttt{CONSIS} \\
    \hline
    \hline
    & & \multicolumn{3}{c|}{\texttt{EN$\rightarrow$DE} \tiny{(newstest2019)}} & \multicolumn{3}{c}{\texttt{DE$\rightarrow$EN} \tiny{(newstest2019)}} \\
    \hline
    & BPE & 39.70\tiny{$\pm$0.71} & -- & -- & 40.01\tiny{$\pm$0.65} & -- & -- \\
    original & BPE-Dropout & 39.65\tiny{$\pm$0.73} & -- & -- & 40.16\tiny{$\pm$0.66} & -- & -- \\
    & SentencePiece & 39.85\tiny{$\pm$0.75} & -- & -- & 40.25\tiny{$\pm$0.67} & -- & -- \\
    \hline
    & BPE & 29.38\tiny{$\pm$0.60} & 74.01\tiny{$\pm$0.95} & 60.59\tiny{$\pm$0.80} & 33.48\tiny{$\pm$0.61} & 83.69\tiny{$\pm$0.96} & 71.51\tiny{$\pm$0.74} \\
    + misspelling & BPE-Dropout & 33.13\tiny{$\pm$0.70} & 83.55\tiny{$\pm$0.92} & 70.74\tiny{$\pm$0.77} & 35.97\tiny{$\pm$0.64} & 89.58\tiny{$\pm$0.78} & 78.33\tiny{$\pm$0.64} \\
    & SentencePiece & 31.87\tiny{$\pm$0.66} & 79.99\tiny{$\pm$0.97} & 66.40\tiny{$\pm$0.76} & 35.26\tiny{$\pm$0.66} & 87.61\tiny{$\pm$0.91} & 74.09\tiny{$\pm$0.74} \\
    \hline
    & BPE & 31.61\tiny{$\pm$0.74} & 79.63\tiny{$\pm$1.31} & 73.26\tiny{$\pm$1.19} & 33.72\tiny{$\pm$0.69} & 84.27\tiny{$\pm$1.15} & 73.19\tiny{$\pm$1.13} \\
    + case-changing & BPE-Dropout & 35.04\tiny{$\pm$0.73} & 88.37\tiny{$\pm$0.97} & 80.04\tiny{$\pm$0.99} & 36.34\tiny{$\pm$0.69} & 90.48\tiny{$\pm$0.95} & 78.96\tiny{$\pm$0.96} \\
    & SentencePiece & 33.49\tiny{$\pm$0.73} & 84.05\tiny{$\pm$1.09} & 76.24\tiny{$\pm$1.09} & 34.48\tiny{$\pm$0.71} & 85.65\tiny{$\pm$1.10} & 74.55\tiny{$\pm$1.10} \\
    \hline
    \hline
    & & \multicolumn{3}{c|}{\texttt{EN$\rightarrow$FR} \tiny{(newstest2014)}} & \multicolumn{3}{c}{\texttt{FR$\rightarrow$EN} \tiny{(newstest2014)}} \\
    \hline
    & BPE & 41.47\tiny{$\pm$0.48} & -- & -- & 39.24\tiny{$\pm$0.50} & -- & -- \\
    original & BPE-Dropout & 40.72\tiny{$\pm$0.48} & -- & -- & 39.22\tiny{$\pm$0.50} & -- & -- \\
    & SentencePiece & 41.05\tiny{$\pm$0.48} & -- & -- & 39.14\tiny{$\pm$0.50} & -- & -- \\
    \hline
    & BPE & 34.01\tiny{$\pm$0.45} & 82.01\tiny{$\pm$0.66} & 71.59\tiny{$\pm$0.53} & 32.62\tiny{$\pm$0.48} & 83.13\tiny{$\pm$0.63} & 73.05\tiny{$\pm$0.49} \\
    + misspelling & BPE-Dropout & 35.98\tiny{$\pm$0.46} & 88.36\tiny{$\pm$0.59} & 78.49\tiny{$\pm$0.48} & 34.71\tiny{$\pm$0.48} & 88.51\tiny{$\pm$0.60} & 79.27\tiny{$\pm$0.50} \\
    & SentencePiece & 34.78\tiny{$\pm$0.45} & 84.72\tiny{$\pm$0.59} & 75.28\tiny{$\pm$0.51} & 33.44\tiny{$\pm$0.48} & 85.43\tiny{$\pm$0.62} & 75.28\tiny{$\pm$0.50} \\
    \hline
    & BPE & 34.75\tiny{$\pm$0.54} & 83.81\tiny{$\pm$0.97} & 79.34\tiny{$\pm$0.93} & 32.31\tiny{$\pm$0.54} & 82.34\tiny{$\pm$0.96} & 76.56\tiny{$\pm$0.95} \\
    + case-changing & BPE-Dropout & 38.28\tiny{$\pm$0.47} & 94.00\tiny{$\pm$0.55} & 86.28\tiny{$\pm$0.58} & 35.78\tiny{$\pm$0.50} & 91.24\tiny{$\pm$0.65} & 84.47\tiny{$\pm$0.65} \\
    & SentencePiece & 36.49\tiny{$\pm$0.50} & 88.87\tiny{$\pm$0.74} & 82.73\tiny{$\pm$0.76} & 33.51\tiny{$\pm$0.54} & 85.61\tiny{$\pm$0.84} & 78.18\tiny{$\pm$0.88} \\
    \hline
    \hline
    & & \multicolumn{3}{c|}{\texttt{EN$\rightarrow$FI} \tiny{(newstest2019)}} & \multicolumn{3}{c}{\texttt{FI$\rightarrow$EN} \tiny{(newstest2019)}} \\
    \hline
    & BPE & 20.43\tiny{$\pm$0.55} & -- & -- & 24.31\tiny{$\pm$0.59} & -- & -- \\
    original & BPE-Dropout & 20.01\tiny{$\pm$0.54} & -- & -- & 24.51\tiny{$\pm$0.57} & -- & -- \\
    & SentencePiece & 20.63\tiny{$\pm$0.57} & -- & -- & 24.67\tiny{$\pm$0.60} & -- & -- \\
    \hline
    & BPE & 15.20\tiny{$\pm$0.46} & 74.42\tiny{$\pm$1.39} & 52.76\tiny{$\pm$0.89} & 21.27\tiny{$\pm$0.54} & 87.47\tiny{$\pm$1.14} & 70.06\tiny{$\pm$0.89} \\
    + misspelling & BPE-Dropout & 17.39\tiny{$\pm$0.50} & 86.95\tiny{$\pm$1.43} & 63.63\tiny{$\pm$0.86} & 22.40\tiny{$\pm$0.55} & 91.38\tiny{$\pm$1.06} & 75.18\tiny{$\pm$0.83} \\
    & SentencePiece & 16.73\tiny{$\pm$0.51} & 81.09\tiny{$\pm$1.52}& 57.45\tiny{$\pm$0.85} & 21.89\tiny{$\pm$0.57} & 88.76\tiny{$\pm$1.19}& 70.57\tiny{$\pm$0.87} \\
    \hline
    & BPE & 15.65\tiny{$\pm$0.53} & 76.63\tiny{$\pm$1.71} & 68.27\tiny{$\pm$1.44} & 20.71\tiny{$\pm$0.58} & 85.20\tiny{$\pm$1.32} & 74.85\tiny{$\pm$1.16} \\
    + case-changing & BPE-Dropout & 17.19\tiny{$\pm$0.53} & 85.92\tiny{$\pm$1.39} & 72.76\tiny{$\pm$1.30} & 23.10\tiny{$\pm$0.58} & 94.26\tiny{$\pm$1.09} & 79.67\tiny{$\pm$1.00} \\
    & SentencePiece & 15.72\tiny{$\pm$0.54} & 76.19\tiny{$\pm$1.72} & 67.73\tiny{$\pm$1.40} & 21.50\tiny{$\pm$0.58} & 87.16\tiny{$\pm$1.26} & 76.29\tiny{$\pm$1.12} \\
    \hline
    \hline
    & & \multicolumn{3}{c|}{\texttt{EN$\rightarrow$JA} \tiny{(KTJ)}} & \multicolumn{3}{c}{\texttt{JA$\rightarrow$EN} \tiny{(KTJ)}} \\
    \hline
    & BPE & 24.28\tiny{$\pm$0.53} & -- & -- & 22.80\tiny{$\pm$0.51} & -- & -- \\
    original & BPE-Dropout & 24.11\tiny{$\pm$0.51} & -- & -- & 22.21\tiny{$\pm$0.52} & -- & -- \\
    & SentencePiece & 22.63\tiny{$\pm$0.45} & -- & -- & 22.99\tiny{$\pm$0.50} & -- & -- \\
    \hline
    & BPE & 19.82\tiny{$\pm$0.47} & 81.66\tiny{$\pm$1.09} & 54.84\tiny{$\pm$0.73} & 18.20\tiny{$\pm$0.45} & 79.83\tiny{$\pm$1.20} & 52.34\tiny{$\pm$0.74} \\
    + misspelling & BPE-Dropout & 22.01\tiny{$\pm$0.49} & 91.30\tiny{$\pm$0.95} & 63.21\tiny{$\pm$0.78} & 18.89\tiny{$\pm$0.47} & 85.06\tiny{$\pm$1.17} & 56.43\tiny{$\pm$0.78} \\
    & SentencePiece & 19.85\tiny{$\pm$0.41} & 87.69\tiny{$\pm$1.05} & 61.25\tiny{$\pm$0.80} & 18.97\tiny{$\pm$0.46} & 82.53\tiny{$\pm$1.15} & 56.40\tiny{$\pm$0.73} \\
    \hline
    & BPE & 20.35\tiny{$\pm$0.51} & 83.83\tiny{$\pm$1.13} & 68.10\tiny{$\pm$1.25} & -- & -- & -- \\
    + case-changing & BPE-Dropout & 21.44\tiny{$\pm$0.49} & 88.91\tiny{$\pm$1.00} & 72.96\tiny{$\pm$1.13} & -- & -- & -- \\
    & SentencePiece & 19.99\tiny{$\pm$0.44} & 88.32\tiny{$\pm$1.06} & 73.52\tiny{$\pm$1.10} & -- & -- & -- \\
\end{tabular}}
\caption{BLEU, robustness (in percentage), and consistency scores of different subword segmentation methods on original and perturbed test sets. We report mean and standard deviation using bootstrap resampling \citep{koehn-2004-statistical}. Subword regularization makes NMT models more robust to input perturbations.}
\label{tab:robustness}
\end{table*}

\begin{table*}[!p]
\centering
\adjustbox{width=0.9\width}{
\begin{tabular}{l|l|r|r|r|r|r}
    & & \multicolumn{4}{c|}{MTNT \tiny{(mtnt2019)}} & \multicolumn{1}{c}{4SQ} \\
    & Model & \texttt{EN$\rightarrow$JA} & \texttt{JA$\rightarrow$EN} & \texttt{EN$\rightarrow$FR} & \texttt{FR$\rightarrow$EN} & \texttt{FR$\rightarrow$EN} \\
    \hline
    & BPE & 10.75\tiny{$\pm$0.49} & 9.68\tiny{$\pm$0.59} & 34.15\tiny{$\pm$0.93} & 45.84\tiny{$\pm$0.89} & 30.96\tiny{$\pm$0.85} \\
    baseline & BPE-Dropout & 10.76\tiny{$\pm$0.47} & 9.26\tiny{$\pm$0.64} & 33.39\tiny{$\pm$0.95} & 45.84\tiny{$\pm$0.90} & 31.28\tiny{$\pm$0.84} \\
    & SentencePiece & 10.52\tiny{$\pm$0.51} & 9.52\tiny{$\pm$0.68} & 33.75\tiny{$\pm$0.91} & 45.94\tiny{$\pm$0.92} & 31.44\tiny{$\pm$0.85} \\
    \hline
    & BPE & 14.88\tiny{$\pm$0.52} & 10.47\tiny{$\pm$0.69} & 35.11\tiny{$\pm$0.95} & 46.49\tiny{$\pm$0.90} & 34.83\tiny{$\pm$0.86} \\
    fine-tuning & BPE-Dropout & 15.26\tiny{$\pm$0.53} & 11.13\tiny{$\pm$0.68} & 34.80\tiny{$\pm$0.93} & 46.88\tiny{$\pm$0.88} & 34.72\tiny{$\pm$0.84} \\
    & SentencePiece & 14.68\tiny{$\pm$0.53} & 11.19\tiny{$\pm$0.72} & 34.71\tiny{$\pm$0.93} & 46.89\tiny{$\pm$0.90} & 34.59\tiny{$\pm$0.86} \\
\end{tabular}}
\caption{BLEU scores of using different subword segmentation methods on two datasets with natural noise. Subword regularization methods do not achieve consistent improvement over BPE, nor with or without fine-tuning. }
\label{tab:mtnt-4sq}
\end{table*}

As shown in Table~\ref{tab:robustness}, there is no clear winner among the three subword segmentation models based on BLEU scores on original WMT or KTJ test sets. This observation is different from results reported by \citet{kudo-2018-subword} and \citet{bpe-dropout}. One major difference from previous work is the size of the training data, which is much larger in our experiments -- subword regularization is presumably preferable on low-resource settings.

However, both our proposed metrics (i.e., robustness and consistency) show clear trends of models' robustness to input perturbations across all languages we tested: BPE-Dropout $>$ SentencePiece $>$ BPE. This suggests that although we did not observe a significant impact of subword regularization on generic translation quality, the robustness of the models is indeed improved drastically.

Unfortunately, it is unclear if subword regularization can help translating real-world noisy input, as shown in Table~\ref{tab:mtnt-4sq}. MTNT and 4SQ contain several natural noise types such as grammar errors, emojis, with misspelling as the dominating noise type for English and French. The training data we use may already cover common natural misspellings, perhaps contributing to the failure of regularization methods to improve over BPE in this case.

\paragraph{Robustness Versus Consistency} Variation in output is not necessarily in itself a marker of reduced translation quality, but empirically, consistency and robustness nearly always provide same model rankings in Table~\ref{tab:robustness}. We conduct more comprehensive analysis on the correlation between them, and we collect additional data points by varying the noise level of both perturbations. Specifically, we use the following word misspelling probabilities: $\{0.05, 0.1, 0.15, 0.2\}$ and the following sentence case-changing probability values: $\{0.3, 0.5, 0.7, 0.9\}$.

\begin{figure}[t]
	\centering
	\includegraphics[width=\columnwidth]{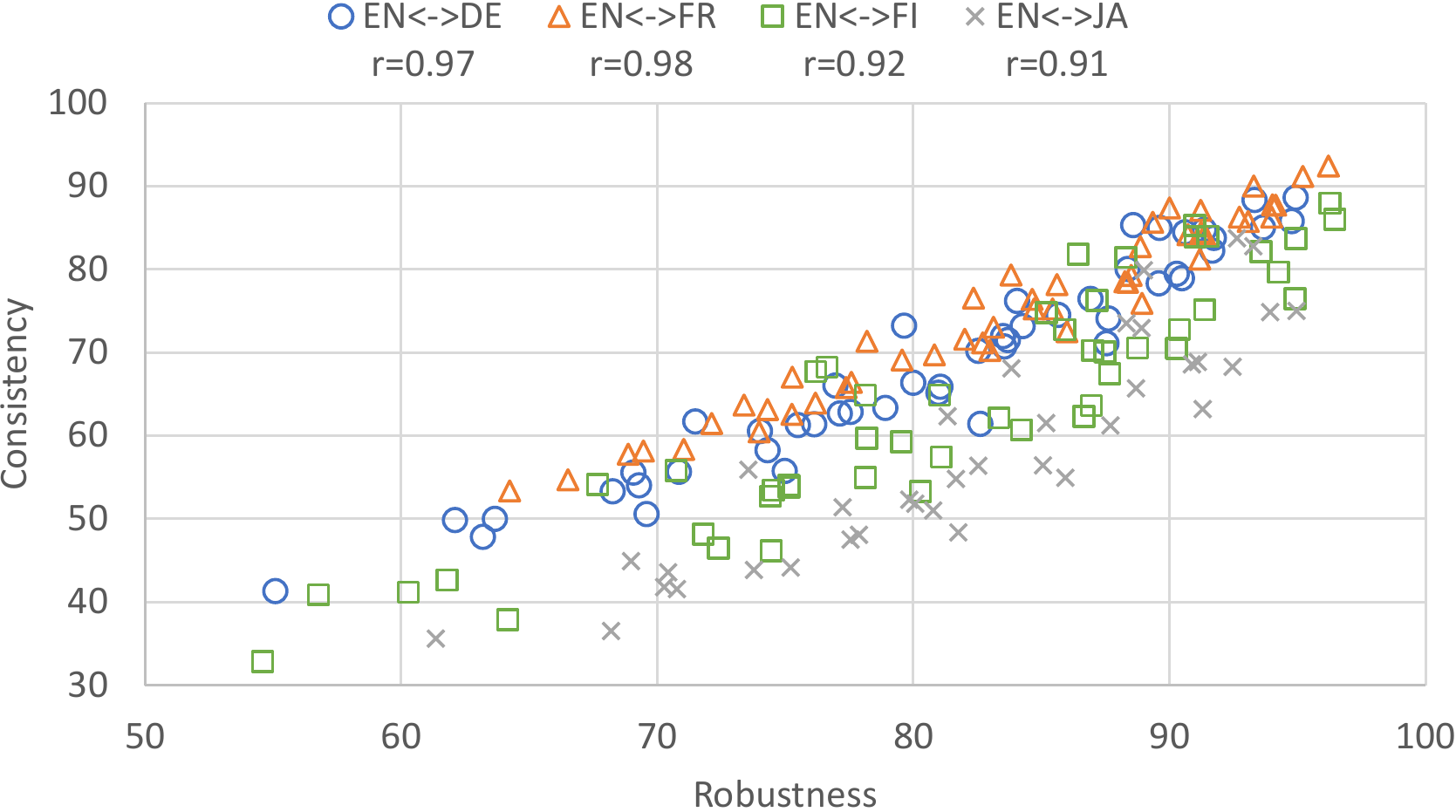}
	\caption{Robustness (in percentage) and consistency are highly correlated within each language pair. Correlation coefficients are marked in the legend.}
	\label{fig:correlation}
\end{figure}

As illustrated in Figure~\ref{fig:correlation}, consistency strongly correlates with robustness (sample Pearson's $r=0.91$ to $0.98$) within each language pair. This suggests that for this class of models, low consistency signals a drop in translation quality and the consistency score can be used as a robustness proxy when the reference translation is unavailable.

\paragraph{Robustness Versus Noise Level}

In this paper, robustness is defined by giving a fixed perturbation function and its noise level. We observe consistent model rankings across language pairs, but is it still true if we vary the noise level?

\begin{figure}[t]
	\centering
	\includegraphics[width=\columnwidth]{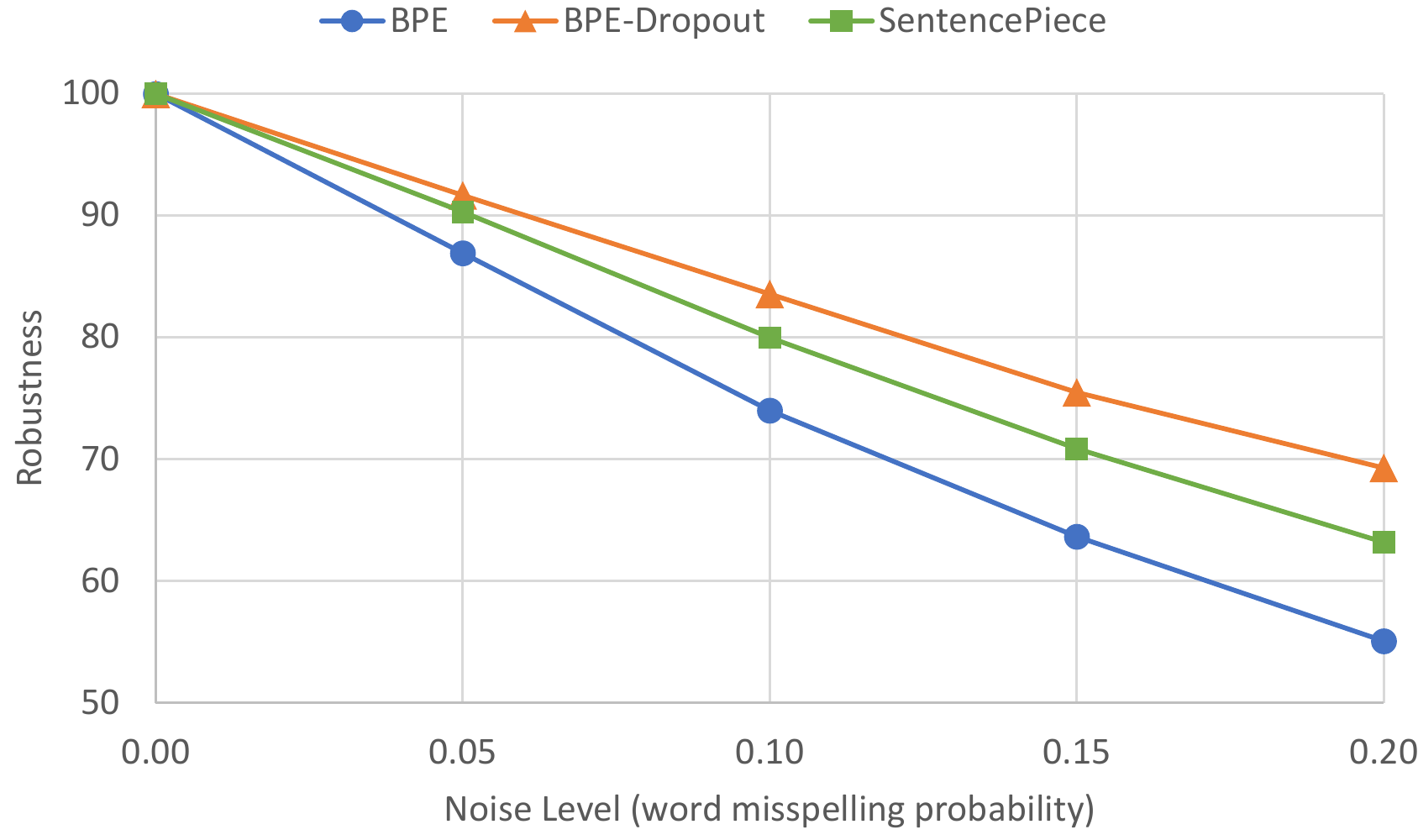}
	\caption{Varying the synthetic word misspelling probability for \texttt{EN$\rightarrow$DE} models does not change the model ranking w.r.t. robustness (in percentage).}
	\label{fig:noise-level}
\end{figure}

To test this, we plot the robustness data points from the last section against the noise level. Focusing on the misspelling perturbation for \texttt{EN$\rightarrow$DE} models, Figure~\ref{fig:noise-level} shows that varying the word misspelling probability does not change the ranking of the models, and the gap in the robustness measurement only increases with larger amount of noise. This observation applies to all perturbations and language pairs we investigated.

\section{Conclusion}
We proposed two additional measures for NMT robustness which can be applied when both original and noisy inputs are available. These measure robustness as relative degradation in quality as well as consistency which quantifies variation in translation output irrespective of reference translations. We also tested two popular subword regularization techniques and their effect on overall performance and robustness. Our robustness metrics reveal a clear trend of subword regularization being much more robust to input perturbations than standard BPE. Furthermore, we identify a strong correlation between robustness and consistency in these models indicating that consistency can be used to estimate robustness on data sets or domains lacking reference translations.

\section{Acknowledgements}
We thank the anonymous reviewers for their comments and suggestions.

\bibliography{biblio}
\bibliographystyle{acl_natbib}

\end{document}